\documentclass[%
 aip,
 jcp,
 amsmath,amssymb,
 reprint,%
]{revtex4-1}

\usepackage{graphicx}
\usepackage{dcolumn}
\usepackage{bm}
\usepackage{xfrac}
\usepackage{textcomp}
\usepackage{xcolor}
\usepackage[inkscapeformat=png]{svg}
\usepackage[unicode=true]{hyperref}
\hypersetup{breaklinks=true, colorlinks=true, urlcolor= blue, citecolor=blue}

\usepackage[utf8]{inputenc}
\usepackage[T1]{fontenc}
\usepackage{mathptmx}
\usepackage{etoolbox}

\makeatletter
\def\@email#1#2{%
 \endgroup
 \patchcmd{\titleblock@produce}
  {\frontmatter@RRAPformat}
  {\frontmatter@RRAPformat{\produce@RRAP{*#1\href{mailto:#2}{#2}}}\frontmatter@RRAPformat}
  {}{}
}%
\makeatother
\begin{document}

\preprint{AIP/123-QED}

\title{Physics-Constrained Neural Surrogate for Domain Growth Prediction in Systems with Conserved Kinetics}

\author{Vijay Yadav}
 \affiliation{Department of Physics, Indian Institute of Technology Jodhpur, Jodhpur, Rajasthan 342 030, India}

\author{Pallvi Pandey}
 \affiliation{Department of Physics, Indian Institute of Technology Jodhpur, Jodhpur, Rajasthan 342 030, India}

\author{Madhu Priya}
\affiliation{Department of Physics, Birla Institute of Technology Mesra, Ranchi, Jharkhand 835 215, India}

\author{Manish Dev Shrimali}
\affiliation{ 
Department of Physics, Central University of Rajasthan, Ajmer, Rajasthan 305 817, India
}

\author{Prabhat K. Jaiswal}%
\email{prabhat.jaiswal@iitj.ac.in}
\affiliation{Department of Physics, Indian Institute of Technology Jodhpur, Jodhpur, Rajasthan 342 030, India}

\date{\today}

\begin{abstract}
The spatiotemporal evolution of many physical, chemical, and biological systems is described by nonlinear partial differential equations (PDEs). Recently, deep neural network–based surrogate models have emerged as efficient alternatives to computationally expensive numerical PDE solvers. In this work, we propose a physics-constrained deep neural network as a surrogate model to learn the microstructural evolution of a binary mixture, in which conservation of the order parameter is imposed directly on the network output as a hard constraint. We train the model to accurately predict the time-evolution of phase separation in binary mixtures governed by the Cahn–Hilliard equation. We show that predictions from our trained surrogate model remain stable and accurate over long-time rollouts for both critical and off-critical mixtures and preserve the mixture composition throughout evolution. In contrast, a variant in which conservation is enforced only via a penalty term in the loss function drifts away from the initial composition and significantly loses predictive accuracy over the same rollout. This establishes that the hard constraint is essential for long-time stability and order parameter conservation. We also show that our model accurately captures the growth of domain size and is consistent with the Lifshitz–Slyozov domain-growth law. These results demonstrate the effectiveness of the proposed framework for modeling systems with conserved kinetics, and the construction extends directly to other systems with conserved quantities.
\end{abstract}

\maketitle

\section{\label{sec:intro} Introduction}
Nonlinear partial differential equations (PDEs) play a fundamental role in modeling and understanding complex dynamical systems across a wide range of scientific and engineering applications~\cite{Hohenberg1977, Evans2010, Cahn1958, Keller1970, Hillen2009}. For example, the Navier–Stokes equations are used to describe the evolution of the velocity field of a fluid, and the Keller–Segel model is used to describe the collective movement and aggregation of cells in response to chemical signals~\cite{Keller1970}. In most cases, obtaining analytical solutions to these nonlinear PDEs is highly challenging, and often impossible~\cite{geng2024acnn}. Consequently, the development of efficient and accurate numerical methods for approximating their solutions becomes essential for simulating and understanding their dynamics. Commonly used numerical techniques rely on spatial and temporal discretization, including finite-difference, finite-element, and spectral methods. However, these techniques often require complex meshing and iterative solutions of large, sparse systems, leading to high computational cost and limited scalability on modern parallel architectures~\cite{chen2021dfsnet, sun2020surrogate, dwivedi2020pielm}.

In recent years, machine learning has been extensively investigated for the study of PDEs and can be broadly classified into two categories: (i) physics-informed methods, which explicitly embed the governing equations into learning, and (ii) purely data-driven surrogate and operator-learning methods, which infer system behavior directly from data. In 2019, Raissi et al. introduced physics-informed neural networks (PINNs), which have achieved notable success in the first category~\cite{raissi2019pinn}. In the standard PINN framework, physical priors, including governing equations, boundary conditions, and initial conditions, are explicitly embedded into the training process~\cite{raissi2019pinn, karniadakis2021piml, chen2021dfsnet}. The core idea is to represent the PDE residual as part of the loss function of a neural network, which is minimized during training to enforce physical consistency. The second category includes data-driven surrogate and operator-learning models, such as convolutional neural networks (CNNs)~\cite{bhatnagar2019cnnflow, kim2019deepfluids, zhu2019pcdl}, recurrent neural networks (RNNs)~\cite{montes2021accelerating, srinivasan2019turbulence, gajamannage2023rnn, wiewel2019latent}, and neural operators (NOs)~\cite{lu2021deeponet, li2021fourier, diab2025tno, fang2025chhs}, which learn the spatiotemporal evolution of the system directly from data without explicitly enforcing the governing equations.

Our work focuses on the Cahn–Hilliard (CH) equation~\cite{Cahn1958, cahn1961spinodal}, which is a fourth-order nonlinear PDE that describes the evolution of a conserved order parameter, typically the concentration field. It plays a fundamental role in modeling phase separation and coarsening phenomena in binary mixtures, alloys, and other multiphase systems. Recently, approaches from both categories discussed above have been explored as efficient alternatives to traditional, computationally expensive numerical solvers for learning CH dynamics. However, its solution exhibits several features that pose challenges for PINNs and neural operator modeling. The presence of fourth-order spatial derivatives increases the complexity of automatic differentiation and often results in unstable gradient propagation during optimization~\cite{wang2021gradient, krishnapriyan2021failure}. Furthermore, the nonlinear free-energy formulation produces thin diffuse interfaces and high-frequency structures, while conventional PINNs tend to favor smooth solutions due to spectral bias~\cite{wang2021ntkpinn}. The phase separation dynamics also involve multiple temporal scales, leading to stiffness that complicates training and sampling strategies. In addition, the conservation of order parameter, an intrinsic property of the CH equation, is not naturally enforced in standard PINN frameworks, which may result in physically inconsistent solutions.

Several improvements, including conservative PINNs (cPINNs)~\cite{jagtap2020} and adaptive progressive marching PINNs (APM-PINNs)~\cite{hu2025}, have been proposed to mitigate these challenges by enforcing conservation laws and improving training stability for stiff dynamics~\cite{qiumei2024, geng2024acnn, kiyani2022ml, geng2025endtoend, he2023adaptivepinn, mattey2022bcpinn}. In parallel, data-driven approaches, such as recurrent neural networks (RNNs), have been explored to accelerate simulations of CH dynamics~\cite{montes2021accelerating}. In addition, our recent studies investigated modern machine learning architectures, including reservoir computing (RC)~\cite{Chauha2023} and graph neural networks (GNNs)~\cite{yadav2025}, to learn the evolution of phase separation in a binary mixture governed by the CH equation. However, these models fail to preserve conservation of the order parameter, particularly during long-term autoregressive rollouts, where small prediction errors accumulate over time. This leads to a gradual loss of the conserved quantity and results in physically inconsistent dynamics, highlighting the need for learning frameworks that explicitly enforce conservation while ensuring stable long-time predictions for CH dynamics.

In this work, we present a physics-constrained convolutional network, a data-driven method to model systems with conserved order parameters. The proposed model is inspired by the modern residual U-Net architecture~\cite{ronneberger_u-net_2015, jha2019resunetpp} and enforces conservation of the order parameter as a hard constraint: an affine projection applied to the network output at every autoregressive step fixes the spatial mean exactly. Periodic boundary conditions are likewise imposed architecturally, through circular convolutions. In addition, a self-attention mechanism is introduced to capture global patterns in the evolving microstructure. These modifications improve long-term rollout accuracy while conserving the order parameter to machine precision throughout the evolution. We demonstrate the effectiveness of the proposed model by evaluating physically relevant quantities, including growth laws, dynamical scaling behavior, order parameter distributions, and conservation properties, rather than relying solely on one-to-one field prediction accuracy.

The paper is organized as follows. Section~\ref {sec:method} reviews phase-field modeling of phase separation in binary mixtures and describes the proposed model architecture, dataset generation, and training procedure. Section~\ref{sec:results} presents the evaluation of the model on training and validation datasets, followed by an analysis of its performance in capturing domain morphology for both critical and off-critical mixtures, and its ability to reproduce domain growth laws. Finally, Sec.~\ref{sec:conc} provides concluding remarks.

\section{\label{sec:method} Methods}
\subsection{Phase-Field Model of Phase Separation in a Binary Mixture}
We consider a binary (AB) alloy system in which hydrodynamic effects are negligible, so that the dynamics are governed primarily by diffusive transport. The system is described within the phase-field framework using a space–time dependent conserved order parameter $\psi(\vec{r}, t)$, which represents the local difference in densities of the two species. Specifically, the order parameter is defined as
\begin{equation}
    \psi(\vec{r},t) = \rho_A(\vec{r}, t) - \rho_B(\vec{r}, t),
    \label{eqn:order-parameter}
\end{equation}
where $\rho_{\alpha} (\vec{r}, t)$ denotes the local number density of species $\alpha \in \{A, B\}$ at position $\vec{r}$ and time $t$. Positive values of $\psi$ correspond to regions enriched in species $A$, whereas negative values indicate $B$-rich regions. Since the total number of particles is conserved, the order parameter is also a conserved quantity. Consequently, the evolution of $\psi$ must satisfy a local conservation law as phase separation proceeds through the diffusion of $A$-rich and $B$-rich species. The temporal evolution of the order parameter is therefore governed by the continuity equation
\begin{equation}
    \frac{\partial}{\partial t} \psi(\vec{r}, t) = -\vec{\nabla} \cdot \vec{J}(\vec{r}, t),
    \label{eqn:continuity}
\end{equation}
where $\vec{J}(\vec{r}, t)$ denotes the diffusive current associated with the transport of the order parameter. Since the current is driven by gradients in the chemical potential arising from concentration fluctuations, it is assumed to follow a linear phenomenological relation of the form
\begin{equation}
    \vec{J}(\vec{r}, t) = -D\vec{\nabla}\mu(\vec{r}, t),
    \label{eqn:current}
\end{equation}
where $D$ is the diffusion coefficient and $\mu(\vec{r}, t)$ represents the local chemical potential. The chemical potential is obtained from the functional derivative of the Helmholtz free-energy functional $F[\psi]$ with respect to the order parameter
\begin{equation}
\mu(\vec{r}, t) = \frac{\delta F[\psi]}{\delta \psi(\vec{r}, t)}.
\label{eqn:chemical-potential}
\end{equation}

Within the phase-field framework, the free-energy functional of the binary mixture is written as
\begin{equation}
F[\psi] = \int d\vec{r} \left[ f(\psi) + \frac{K}{2} |\vec{\nabla} \psi|^2 \right],
\label{eqn:free-energy}
\end{equation}
where $K$ is the gradient energy coefficient that accounts for the energetic cost associated with spatial variations of the order parameter. For a symmetric binary mixture, the bulk free-energy density is commonly represented by a $\psi^4$ Landau expansion
\begin{equation}
f(\psi) = -\frac{a}{2}(T_c - T) \psi^2 + \frac{b}{4}\psi^4,
\label{eqn:local-free-energy}
\end{equation}
where $a$ and $b$ are positive phenomenological constants, $T$ denotes the system temperature, and $T_c$ is the critical temperature of the binary mixture.

Combining the above relations leads to the Cahn–Hilliard (CH) equation governing the dynamics of the conserved order parameter,
\begin{equation}
\frac{\partial}{\partial t} \psi(\vec{r, t}) = \vec{\nabla} \cdot \left[ D \vec{\nabla} \left( -a(T_c-T)\psi + b\psi^3 - K\nabla^2\psi \right) \right].
\label{eqn:ch}
\end{equation}

After appropriate rescaling of space, time, and the order parameter, the dimensionless Cahn–Hilliard equation can be written as

\begin{equation}
\frac{\partial}{\partial t} \psi(\vec{r, t}) = \vec{\nabla} \cdot \left[ \vec{\nabla} \left( -\psi + \psi^3 - \nabla^2 \psi \right) \right].
\label{eqn:ch-dimensionless}
\end{equation}

Equation~(\ref{eqn:ch-dimensionless}) is solved numerically using an explicit time-stepping scheme on a two–dimensional square lattice. Let $\psi_{i,j}^{t}$ denote the value of the order parameter at lattice site $(i,j)$ at time $t$. Spatial derivatives are approximated using finite differences on a uniform grid with spacing $\Delta x=\Delta y=h$. The discrete Laplacian of the order parameter is given by

\begin{equation}
\nabla_h^2 \psi_{i,j}^{t} = \frac{ \psi_{i+1,j}^{t} + \psi_{i-1,j}^{t} + \psi_{i,j+1}^{t} + \psi_{i,j-1}^{t} - 4\psi_{i,j}^{t} }{h^2}.
\label{eqn:laplacian}
\end{equation}

Using this expression, the chemical potential at each lattice site is computed as

\begin{equation}
\mu_{i,j}^{t} = -\psi_{i,j}^{t} + (\psi_{i,j}^{t})^{3} - \nabla^2_h \psi_{i,j}^{t}.
\label{eqn:discrete_mu}
\end{equation}

The time evolution of the order parameter field is then updated according to
\begin{equation}
\psi_{i,j}^{t+\Delta t} = \psi_{i,j}^{t} + \Delta t \, \nabla_h^2 \mu_{i,j}^{t},
\label{eqn:time-stepping}
\end{equation}
where $\nabla_h^2$ denotes the discrete Laplacian operator. In the simulations, the spatial discretization is chosen as $\Delta x=\Delta y=1$ ($h=1$) and the time step is $\Delta t=0.01$. Periodic boundary conditions are imposed in both spatial directions to ensure conservation of the order parameter throughout the evolution.

The evolution of the order parameter leads to the emergence and coarsening of domains whose morphology depends sensitively on the initial composition. For a critical mixture, the system develops a bicontinuous morphology characterized by interconnected domains of both phases, reflecting the symmetry of the composition. In contrast, for off-critical mixtures, the system exhibits a droplet morphology in which the minority phase forms isolated domains dispersed within the continuous majority phase.

\begin{figure*}
    \centering
    \includegraphics[width=\textwidth]{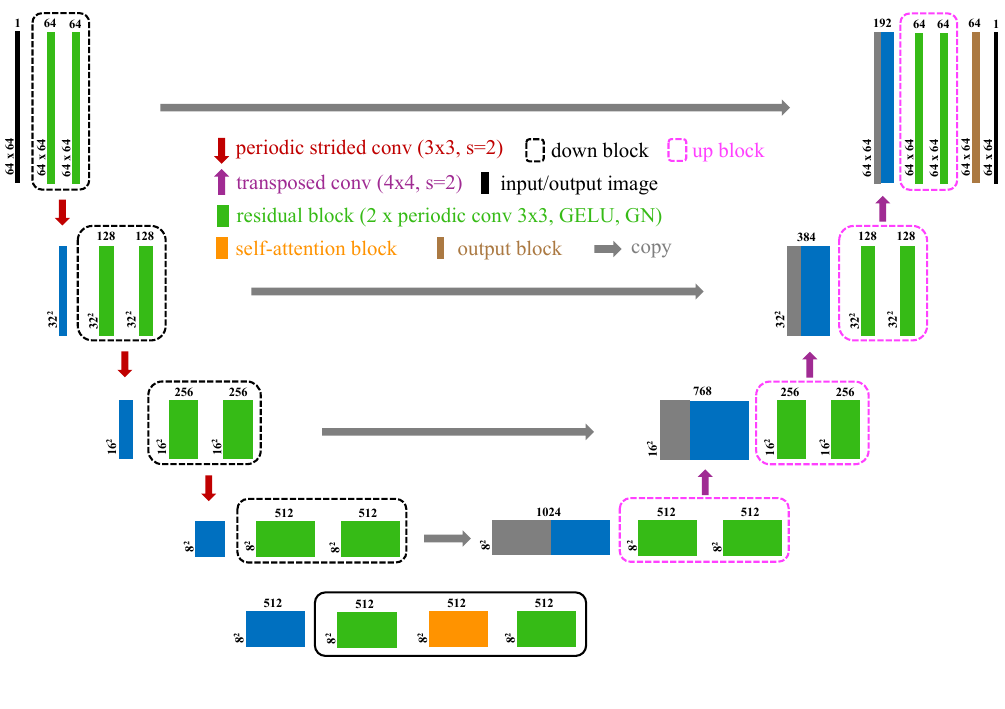}
    \caption{\label{fig:unet_schema} Schematics of Residual U-Net architecture consisting of a contracting path, a central bridge, and an expansive path.}
\end{figure*}

Coarsening is characterized by a time-dependent length scale $L(t)$, the typical domain size~\cite{puri_kinetics_2009, onuki, bray}. For diffusive dynamics such as the Cahn--Hilliard equation, domains grow as $L(t) \sim t^{1/3}$, the Lifshitz--Slyozov law. When a single length scale dominates, the morphology is statistically self-similar, and the equal-time two-point correlation function takes the scaling form 
\begin{align} \label{eq:correlation}
C(\vec{r}, t) &= \frac{1}{V} \int d\vec{R} \Big[
\langle \psi(\vec{R}, t)\psi(\vec{R} + \vec{r}, t) \rangle \nonumber \\
&\quad - \langle \psi(\vec{R}, t) \rangle
\langle \psi(\vec{R} + \vec{r}, t) \rangle
\Big] = s\!\left(\frac{r}{L}\right),
\end{align}
where $V$ is the system volume, $\langle \cdot \rangle$ denotes averaging over independent initial conditions, and $s(x)$ is a time-independent scaling function. Correlation functions at different times therefore collapse onto a single curve when plotted against $r/L(t)$. We take $L(t)$ to be the first zero-crossing of $C(r,t)$.

\subsection{Residual U-Net with Physics Constraint}
To predict the spatiotemporal evolution of phase separation of a binary mixture, we employed a composition-conserving residual U-Net network $f_\theta$, which maps the order-parameter field at time $t$ to its state at $t+100\Delta t$. The network architecture combines deep residual networks and U-Net architecture while enforcing the conservation law as a hard constraint on the network output. The network follows a U-Net--like architecture, consisting of a contracting path, a central bridge, and an expansive path~\cite{ronneberger_u-net_2015}. The input field is first mapped to 64 feature channels by a $3\times3$ convolution with circular padding. The contracting path is composed of four downblocks, while the expansive path symmetrically comprises four upblocks. Each downblock and upblock contains two residual blocks, whereas the central bridge is composed of two residual blocks augmented with a self-attention block. The self-attention block is placed in the central bridge at the lowest spatial resolution, where the receptive field is at its maximum. 

Self-attention~\cite{vaswani_2017} computes, for each spatial position, a weighted average of features at all positions, with weights determined by the similarity of learned query and key projections of the features themselves,
\begin{equation}
\mathrm{Attn}(Q,K,V) = \mathrm{softmax}\!\left(\frac{QK^{\mathsf T}}{\sqrt{d}}\right)V ,
\end{equation}
where $Q$, $K$, and $V$ are linear projections of the input feature map and $d$ is the feature dimension per head; we use four heads. Unlike a convolution, whose support is fixed and local, this coupling is global and depends on the field configuration itself. At the coarsest scale, the feature map is $8\times8$, so the operation couples all 64 positions directly at negligible cost, while spanning the full $64\times64$ domain.

\begin{figure}[!htbp]
    \centering
    \includegraphics[width=0.35\textwidth]{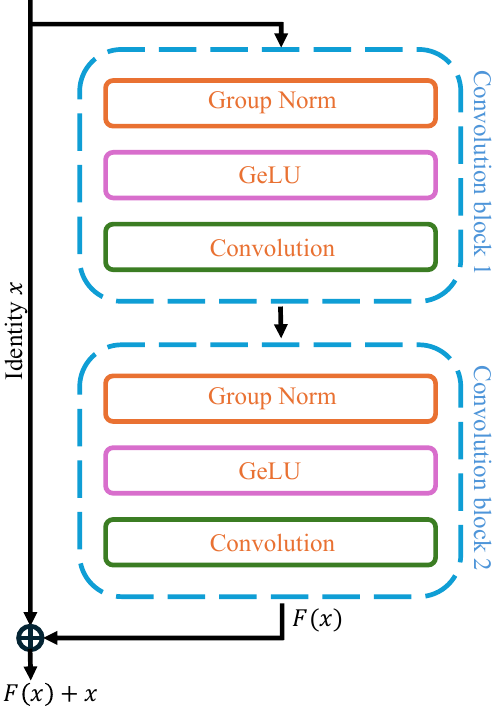}
    \caption{\label{fig:res_block} Schematics of residual block.}
\end{figure}

The network architecture is depicted in Fig.~\ref{fig:unet_schema}. Figure~\ref{fig:res_block} depicts the schematic of residual blocks, which consists of two $3\times3$ convolution blocks and identity mapping, or a $1\times1$ convolution with circular padding when the channel dimension differs. The convolution blocks include group normalization (GN), a GeLU activation, and a convolution layer with circular padding. Circular padding is used throughout so that the network respects the same periodic boundary conditions imposed on the solver in Sec.~\ref{sec:method} A.

After each downblock, downsampling is performed using a $3\times3$ convolution with a stride of 2, which reduces the feature map size by half. Further, we double the number of feature channels at each successive downblock. Downsampling is skipped after the final downblock, and its output is directly passed to the central bridge. Correspondingly, in the expansive path, upsampling is performed using transposed convolutions with a kernel size of 4 and a stride of 2 after each upblock, and is likewise skipped after the final upblock. Upsampling increases the spatial resolution by a factor of two, while the number of feature channels is halved by the succeeding upblock. At each level, the features are concatenated with the corresponding skip connections from the contracting path before being processed by the upblock. After the final upblock, an output layer consisting of group normalization (GN) and a GeLU activation, followed by a $3\times3$ convolutional layer with circular padding, is used to project the multichannel feature map onto the original input space. The network uses a base width of 64 with channel multipliers $(1,2,4,8)$, 8 groups in each normalization layer, and 4 attention heads, giving $4.52\times10^{7}$ trainable parameters.

The conservation of the order parameter is imposed directly on the network output by an affine projection, applied before the training objective is evaluated. Denoting by $\langle\cdot\rangle$ the spatial mean over the lattice, the prediction is obtained as
\begin{equation}
  \hat{\psi}^{\,t+100\Delta t}
  = f_\theta(\psi^{t}) - \bigl\langle f_\theta(\psi^{t})\bigr\rangle
  + \bigl\langle \psi^{t}\bigr\rangle ,
  \label{eq:proj}
\end{equation}
so that $\langle\hat{\psi}^{\,t+100\Delta t}\rangle=\langle\psi^{t}\rangle$ identically, to machine precision, for any $\theta$. The correction in Eq.~\eqref{eq:proj} adds the same constant to every lattice site. It therefore fixes the spatial mean of the prediction without altering any spatial variation in it: interfaces, gradients, and domain morphology are left exactly as the network produced them. It is also the smallest change that achieves the constraint~\cite{baez2024guaranteeing, baez2025guaranteeing}, since any other correction with the same mean would additionally modify the field's structure. The reference mean is taken from the input field and never from the target, so the constraint uses no information unavailable at prediction time and is applied unchanged at every step of an autoregressive rollout. Since the finite-difference solver conserves $\sum_{i,j}\psi_{i,j}$ exactly by virtue of its divergence form, Eq.~\eqref{eq:proj} makes the surrogate reproduce a structural property of the dynamics rather than approximate it.

\subsection{Dataset Generation and Network Training}

We created the dataset by numerically solving the CH equation using a finite-difference method (FDM). For training and validation, we used the system size of $64\times64$, $\Delta x = 1.0$, $\Delta t = 0.01$. Each trajectory is initialized from a homogeneous state with small random fluctuations, $\psi(\mathbf{r},0)$ drawn from a uniform distribution on $[-0.1,0.1]$ and then shifted so that its spatial mean equals the prescribed composition $\psi_0$ exactly. A total of 20 independent trajectories over the interval $t\in[0,300]$ are used from critical ($\psi_0 = 0$) and off-critical ($\psi_0 = -0.4$, $-0.2$, $0.2$, and $0.4$) mixtures for training, four for each composition, and five such independent trajectories, one per composition, for validation. Training and validation trajectories are generated from disjoint sets of random seeds, so that the split is by trajectory rather than by time. For each independent trajectory, we labeled snapshots at any time $t$ as the input image and corresponding snapshots at time $t+100\Delta t$ as the target image. Successive input--target pairs are taken every $10\Delta t$ along a trajectory, so that each trajectory yields 2990 pairs. Therefore, we used 59800 images for training and 14950 images for validation. Network performance is tested on ensembles of independent unseen trajectories generated from FDM. The test set includes the composition $\psi_0 = -0.3$, which is absent from the training data, and rollouts extending well beyond the $t\in[0,300]$ window used for training.

To train the network, we employ a composite loss that balances data fidelity with a physically motivated conservation constraint. Let $y\in\mathbb{R}^{H\times W}$ denote the ground-truth field and $\hat{y}$ the network prediction, where $\hat{y}$ is the projected output of Eq.~\eqref{eq:proj}. Here, we use mean-squared error (MSE)
\begin{equation}
  L_{\mathrm{MSE}} = \frac{1}{N}\sum_{n=1}^{N}\bigl(\hat{y}_n - y_n\bigr)^2 ,
\end{equation}
where $N = HW$ is the number of lattice sites and the average is taken over the samples of a minibatch, to penalize pointwise deviations between predictions and targets. Further, to encode the conservation of the order parameter, which is intrinsic to the underlying conserved dynamics, we introduce an additional penalty term. The spatial mean of each sample is computed as
\begin{equation}
  \langle y\rangle = \frac{1}{HW}\sum_{i,j} y_{ij}, \qquad
  \langle \hat{y}\rangle = \frac{1}{HW}\sum_{i,j} \hat{y}_{ij},
\end{equation}
and deviations between predicted and true means are penalized via
\begin{equation}
  L_{\langle\psi\rangle} = \bigl(\langle\hat{y}\rangle - \langle y\rangle\bigr)^2 .
\end{equation}
The final training objective is given by
\begin{equation}
  L = L_{\mathrm{MSE}} + \alpha\,L_{\langle\psi\rangle} ,
  \label{eq:loss}
\end{equation}
where $\alpha \ge 0$ weights the conservation term. Because the projection of Eq.~\eqref{eq:proj} is applied to the network output before $L$ is evaluated, $\langle\hat{y}\rangle$ already equals the composition of the input field, and $L_{\langle\psi\rangle}$ is satisfied to machine precision throughout training; it is retained as a diagnostic of the constraint. The weights of the network are updated by minimizing the loss function $L$.

The network is trained using PyTorch Lightning with the AdamW optimizer (weight decay $10^{-2}$). We minimized the loss function described in Eq.~\eqref{eq:loss} between true and predicted images. It is worth noting that no data augmentation is employed during training. The model is trained for 20 epochs using a minibatch size of 16 on an NVIDIA A100 GPU with 40 GB of memory. A fixed learning rate of $10^{-4}$ is used throughout training, with the physics regularization parameter set to $\alpha = 1.0$. All computations are carried out in double precision, which is required because the residual composition drift reported below lies at the level of machine precision and would be unresolvable in single precision. Random seeds are fixed, and deterministic algorithms are enabled throughout, so that the reported results are reproducible on identical hardware. During training, we monitored the loss function and the coefficient of determination ($R^2$) to assess convergence and predictive performance. The $R^2$ score quantifies the fraction of variance in the ground-truth data explained by the model, with values closer to unity indicating better agreement.

\section{\label{sec:results} Results and Discussion}
\subsection{Training and Validation}
\begin{figure}[!htbp]
    \centering
    \includegraphics[width=0.35\textwidth]{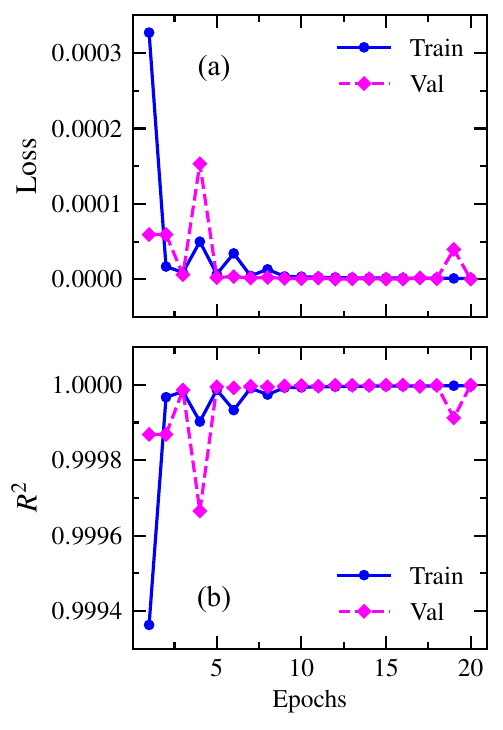}
    \caption{\label{fig:training_metrics} Training and validation performance of the proposed model: (a) loss and (b) coefficient of determination ($R^2$) as a function of training epochs.}
\end{figure}

We first present the training and validation performance of the proposed model in Fig.~\ref{fig:training_metrics}. Figure~\ref{fig:training_metrics}(a) shows the evolution of the loss as a function of training epochs, while Fig.~\ref{fig:training_metrics}(b) illustrates the corresponding behavior of the coefficient of determination ($R^2$). Both metrics exhibit stable convergence, with optimal performance reached at approximately epoch 15 for both training and validation. The model corresponding to this minimum in validation loss is therefore selected for inference, and all results reported in this work are obtained using this checkpoint.

\begin{figure}[!htbp]
    \centering
    \includegraphics[width=0.48\textwidth]{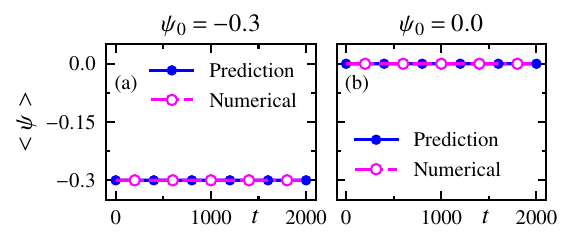}
    \caption{\label{fig:avg_psi} Evolution of the average order parameter, $\langle \psi \rangle$, with time for (a) off-critical composition ($\psi_0 = -0.3$) and (b) critical composition ($\psi_0 = 0.0$). Model predictions are compared with numerical simulations.}
\end{figure}
To assess the stability of the model and its ability to preserve the order parameter conservation, we generated 50 independent trajectories each for critical ($\psi_0 = 0.0$) and off-critical ($\psi_0 = -0.3$) compositions. For each case, the model predicts the full temporal evolution starting from homogeneous initial conditions. The evolution of the average order parameter $\langle \psi \rangle$ is shown in Fig.~\ref{fig:avg_psi}.  $\langle \psi \rangle$ denotes the order parameter averaged over space as well as $50$ independent runs. As seen, the average composition remains effectively constant over time for both off-critical [see Fig.~\ref{fig:avg_psi}(a)] and critical [see Fig.~\ref{fig:avg_psi}(b)] mixtures, demonstrating that the model accurately preserves the global conservation law.

\subsection{Domain Morphology in Critical and Off-Critical Mixtures}
\begin{figure*}
    \includegraphics[width=0.80\textwidth]{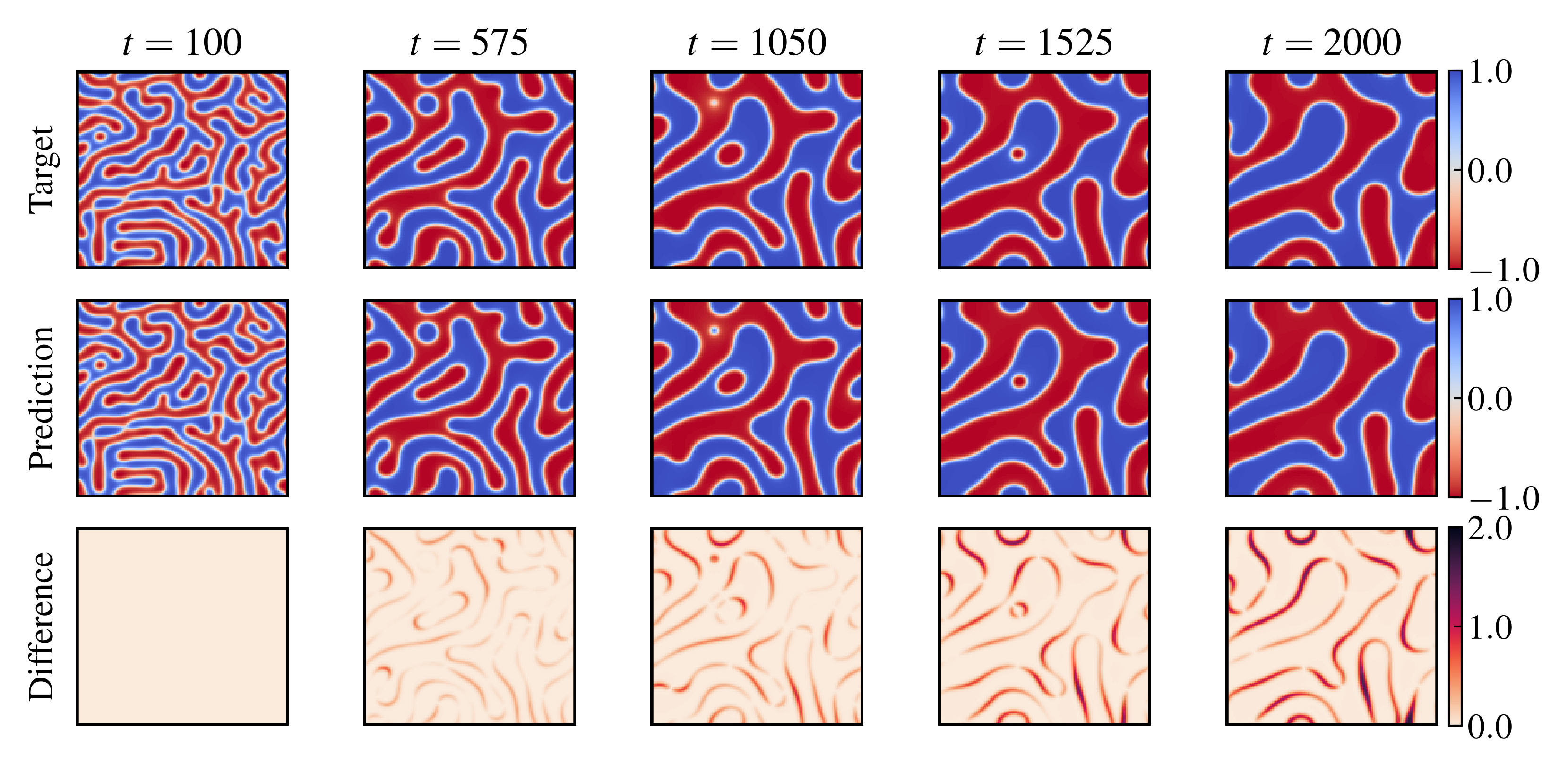}
    \caption{\label{fig:snap_critical} Predictions of phase separation in a homogeneous binary (AB) mixture at critical composition. The top row shows reference solutions obtained by numerically solving the dimensionless Cahn–Hilliard (CH) equation. The middle row presents predictions from the proposed model, while the bottom row depicts the absolute error between the predictions and the reference solutions. All results correspond to a system size of $128 \times 128$, with spatial and temporal resolutions $\Delta x=1.0$ and $\Delta t=0.01$, respectively. Predictions are initiated at $t=100$ to avoid large fluctuations at early times.}
\end{figure*}
To evaluate the performance of our network architecture, we compare the predicted domain morphologies against numerically generated ground-truth solutions. Figure~\ref{fig:snap_critical} illustrates the temporal evolution of a binary mixture at critical composition for system size $128 \times 128$. The top row displays target snapshots obtained via FDM, as discussed in Sec.~\ref{sec:method}. The middle row shows the network predictions generated autoregressively from the same initial condition as the FDM. Finally, the bottom row highlights the absolute difference between the target and the network predictions. In the visualization, the blue (gray) regions correspond to A-rich domains ($\psi > 0$), whereas the red (dark) regions represent B-rich domains ($\psi < 0$). The domain boundaries are represented by the white region where the values of the order parameter are near zero. For this analysis, the prediction is initiated at $t=100$ rather than $t=0$ to avoid early-time fluctuations, and subsequent snapshots are generated autoregressively. The result presented in Fig.~\ref{fig:snap_critical} shows good agreement between target and model predictions for a very long time horizon. It is evident that the model captures domain coarsening accurately, while preserving the overall morphology and connectivity of the domains. Furthermore, the analysis of the difference panel reveals that the error made by our proposed framework occurs primarily at domain boundaries and accumulates over the rollout. However, this accumulation remains confined to the interfaces and grows very slowly. Therefore, the model remains stable and reliable over long rollouts and can be used for faithful simulation of the CH equation.
\begin{figure*}
    \includegraphics[width=0.80\textwidth]{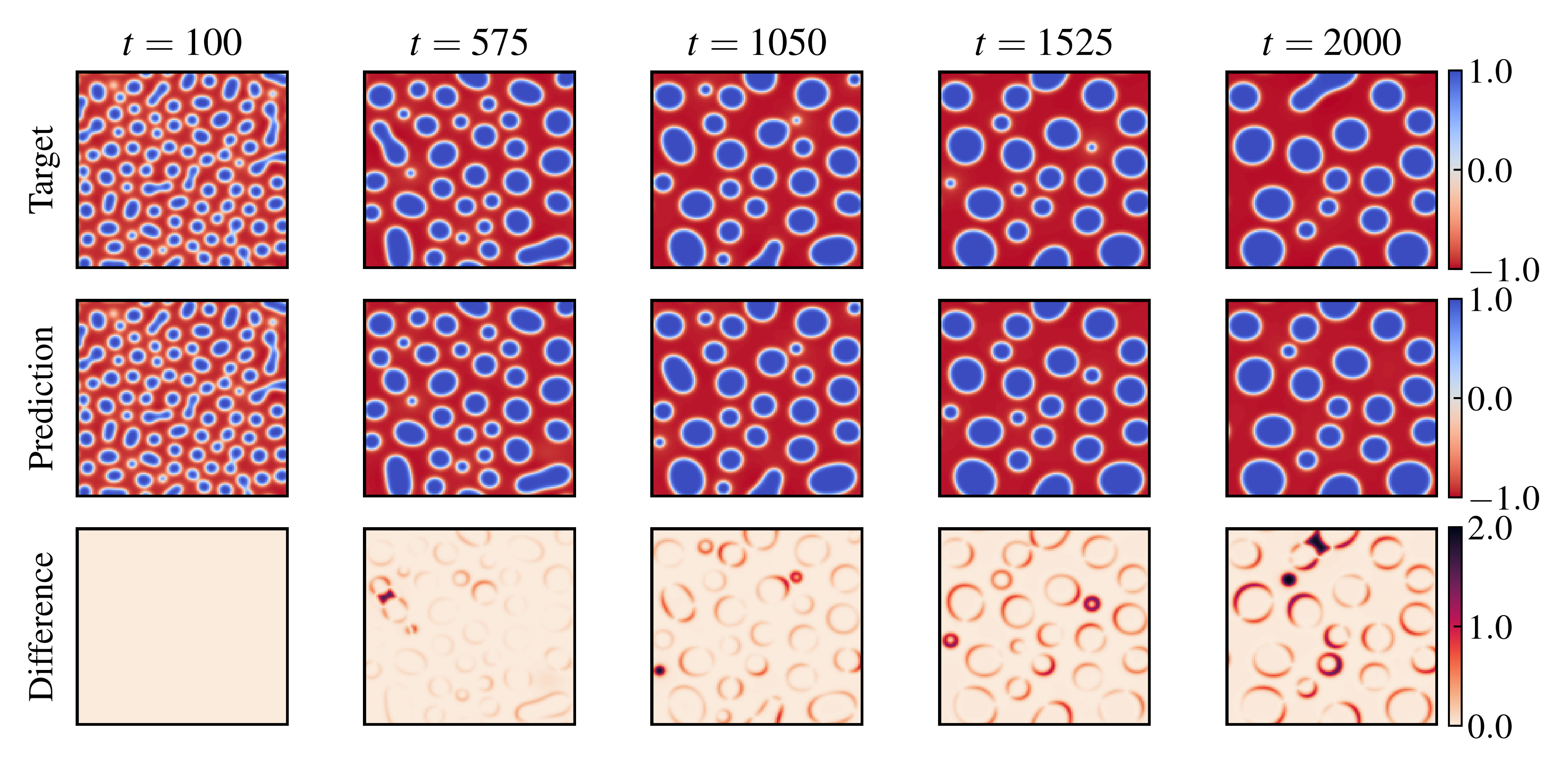}
    \caption{\label{fig:snap_off_critical}The figure layout is the same as in Fig.~\ref{fig:snap_critical}, but for an off-critical mixture. The initial condition is initialized with small-amplitude fluctuations about $\psi_0=-0.3$, corresponding to a $35\%$ A and $65\%$ B mixture.}
\end{figure*}

Next, we present the model predictions for off-critical mixtures, as shown in Fig.~\ref{fig:snap_off_critical}, which depicts the temporal evolution of the domain morphology for a composition of $\psi = -0.3$. Analysis of snapshots at different times shows that the model prediction closely matches the reference solution. Furthermore, the difference panel of Fig.~\ref{fig:snap_off_critical} reveals that the error is no longer confined to the domain boundaries and propagates into the bulk as well. This occurs because the off-critical mixture exhibits a distinct droplet morphology in contrast to the bicontinuous morphology in the critical mixture. These deviations are mainly concentrated in regions undergoing rapid, nonlinear morphological transitions, particularly during active droplet coalescence or breakup events. Nevertheless, the model captures domain coarsening, conservation of $\langle \psi \rangle$, and the predictions remain stable and reliable over very long time scales.
\begin{figure}[!htbp]
    \centering
    \includegraphics[width=0.35\textwidth]{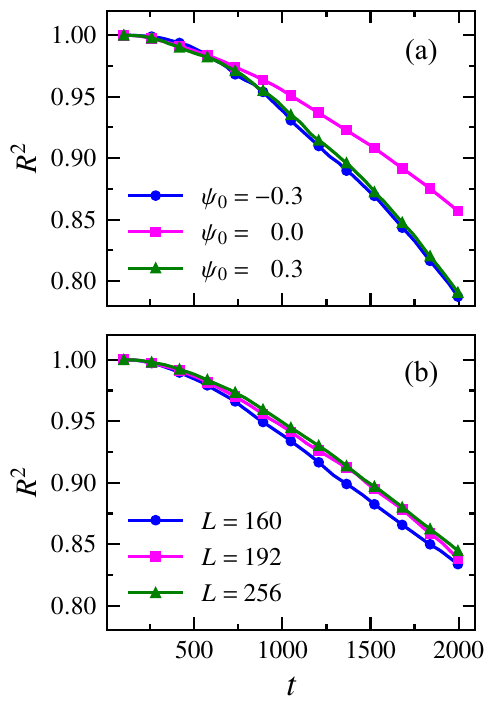}
    \caption{\label{fig:eval_metrics} Temporal evolution of model prediction accuracy measured by $R^2$ for (a) different mixture compositions and (b) different system sizes.}
\end{figure}

To quantify the morphological observations discussed above, Fig.~\ref{fig:eval_metrics} summarizes the predictive performance of the model across different mixture compositions and system sizes. Figure~\ref{fig:eval_metrics}(a) tracks the temporal evolution of the $R^2$ for initial compositions ($\psi_0 = -0.3$, $0.0$, and $0.3$), averaged over $50$ unseen initial conditions. Consistent with our visual assessments, the model maintains excellent accuracy at early times across all compositions, with $R^2 \approx 1$. As the simulation progresses, a gradual decay in accuracy emerges due to the natural accumulation of errors inherent in recursive time-stepping; however, the overall predictive performance remains highly robust. Comparing the different compositions, the network exhibits the highest fidelity for the critical mixture ($\psi_0 = 0.0$), consistently yielding higher $R^2$ values throughout the simulation window. In contrast, the off-critical mixtures ($\psi_0 = \pm 0.3$) display a slightly accelerated degradation in accuracy over time. This quantitative trend aligns with our earlier morphological analysis, where the off-critical droplet morphology involves localized interface motion, coalescence, and breakup events. These processes are more difficult for the model to capture over long trajectories. Further, we evaluated the model performance across different system sizes. We plot $R^2$ as a function of time for system sizes, $L = 160$, $192$, and $256$ in Figure~\ref{fig:eval_metrics}(b) for the critical composition. It is evident that the model performance remains consistent across all system sizes. This is crucial, as although the model was trained on smaller systems, it can accurately predict larger systems without any noticeable degradation in accuracy.

\begin{figure}[!htbp]
    \centering
    \includegraphics[width=0.48\textwidth]{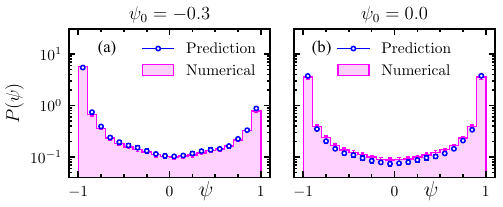}
    \caption{\label{fig:hist} Probability distribution of the order parameter $\psi$ at time $t=2000$ for mixture compositions of (a) $\psi_0 = -0.3$ and (b) $\psi_0 = 0.0$. Results from the numerical solution are shown in magenta, while machine-learning predictions are shown in blue.}
\end{figure}

To further evaluate the model's performance, we examine the distribution of the order parameter. Figure~\ref{fig:hist} shows the probability distribution $P(\psi)$ of the final predicted field after $1900$ autoregressive steps, averaged over $50$ independent realizations, for off-critical ($\psi_0=-0.3$) and critical ($\psi_0=0.0$) compositions. For the off-critical mixture [see Fig.~\ref{fig:hist}(a)], $P(\psi)$ is asymmetric, with a dominant peak near $\psi\approx-1$ corresponding to the majority phase and a smaller peak near $\psi\approx+1$ representing the minority phase. These peaks are separated by a shallow interfacial plateau. The network reproduces this bimodal structure over nearly two decades in $P(\psi)$. The main difference is a slightly higher probability in the intermediate range $|\psi|\lesssim0.9$, accompanied by a slightly lower probability near the majority peak. This suggests marginally broader interfaces and less saturated bulk regions. For the critical mixture [shown in Fig.~\ref{fig:hist}(b)], the distribution is symmetric about $\psi=0$, with equal-height peaks near $\psi\approx\pm1$. The prediction closely matches the numerical reference and reproduces the expected symmetry between the two phases. Overall, the agreement in $P(\psi)$ shows that the network captures not only the spatial morphology and temporal dynamics, but also the relative partitioning between bulk phases and interfaces during long autoregressive rollouts.

\begin{figure}[!htbp]
    \centering
    \includegraphics[width=0.48\textwidth]{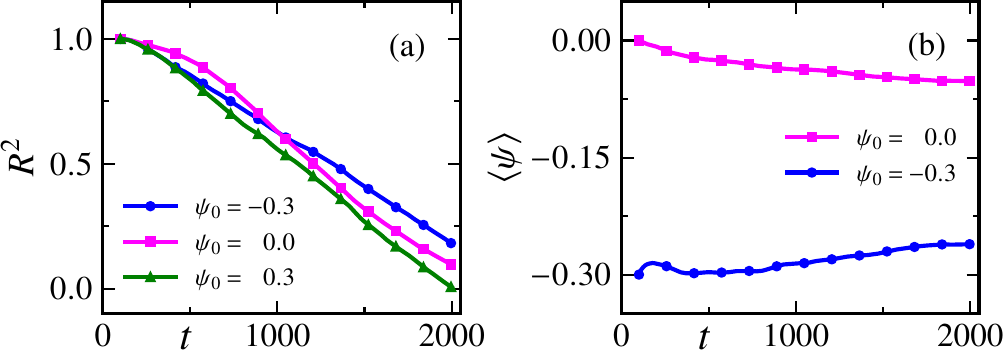}
    \caption{\label{fig:without_constraint} Model performance when conservation is imposed only through the loss function, with the projection removed during both training and rollout. (a) $R^{2}$ as a function of time for initial compositions $\psi_{0} = -0.3$, $0.0$, and $0.3$. (b) Evolution of the average order parameter $\langle\psi\rangle$ for $\psi_{0} = 0.0$ and $\psi_{0} = -0.3$.}
\end{figure}

Finally, we analyze the surrogate's performance when the hard constraint is not applied to the network output. To achieve this, we retrain the network with the unprojected output $\hat{\psi}^{\,t+100\Delta t} =
f_{\theta}(\psi^{t})$ in place of Eq.~\eqref{eq:proj}, so that conservation is enforced only through the penalty term $L_{\langle\psi\rangle}$ of Eq.~\eqref{eq:loss}, which is now active rather than identically zero. The projection is likewise omitted during the rollout. The result is represented in Fig.~\ref{fig:without_constraint}. We find that $R^{2}$ quickly decays and reaches near zero, as shown in Fig.~\ref{fig:without_constraint}(a). This is in sharp contrast to the case where the model output was projected, for which $R^{2}$ remained above $0.8$ after the same rollout time. Figure~\ref{fig:without_constraint}(b) reveals that the average order parameter drifts steadily from its initial value by $\approx \pm 0.05$ at $t = 2000$ for both compositions shown. At $t=2000$, the compositions have shifted from the initial $50:50$ and $35:65$ binary mixtures ($A:B$) to approximately $47.5:52.5$ and $37.5:62.5$, respectively. The composition no longer remains constant, which is not physical for a binary mixture. This reveals an important fact: imposing conservation through the loss function alone cannot guarantee conservation during long rollouts, whereas the projection of Eq.~\eqref{eq:proj} enforces it exactly at every step.

\subsection{Domain growth in Binary Alloy}
\begin{figure}[!htbp]
    \centering
    \includegraphics[width=0.46\textwidth]{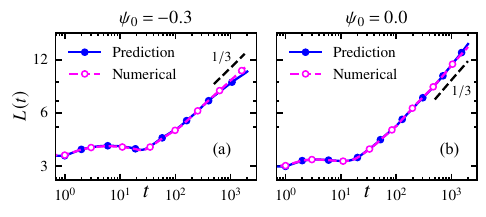}
    \caption{\label{fig:ls_init} Domain growth law for phase separation in a binary mixture for (a) $\psi_0 = -0.3$ and (b) $\psi_0 = 0.0$. The dashed magenta lines represent numerical solver results, and dashed black lines indicate the Lifshitz–Slyozov growth law.}
\end{figure}
The evolution of the characteristic domain length scale $L(t)$ during phase separation is shown in Fig.~\ref{fig:ls_init} for different mixture compositions. The results obtained from the machine-learning model are compared with those from the numerical solver, along with the expected Lifshitz–Slyozov (LS) growth law for reference. Here, predictions were initiated from homogeneous initial conditions. As seen, the predicted growth curves closely follow the numerical results over a wide temporal range for all compositions considered. At early times, the growth of $L(t)$ is relatively slow due to the presence of strong fluctuations and the initial formation of domains. Despite this, the model accurately captures the initial evolution, indicating that it successfully learns the underlying dynamics even in this highly transient regime. As time progresses, the system enters the coarsening regime, where domains grow steadily, and the agreement between prediction and numerical results remains strong.

In the off-critical case ($\psi_0 = - 0.3$) [see Fig.~\ref{fig:ls_init}(a)], a slight deviation between prediction and numerical results emerges at late times, where the predicted growth is marginally slower. This can be attributed to the increased complexity of droplet-based coarsening dynamics, which involve localized interactions such as coalescence and mass transfer between domains. Importantly, the predicted growth behavior remains consistent with the expected LS scaling, as indicated by the dashed reference line in Fig.~\ref{fig:ls_init}(a). For the critical composition ($\psi_0 = 0.0$) [shown in Fig.~\ref{fig:ls_init}(b)], the agreement is nearly indistinguishable throughout the evolution, demonstrating the model’s ability to capture the dynamics of bicontinuous structures with high fidelity. This demonstrates that the model not only reproduces the instantaneous morphology but also captures the correct temporal evolution of the characteristic length scale governing the coarsening process.

\begin{figure}[!htbp]
    \centering
    \includegraphics[width=0.48\textwidth]{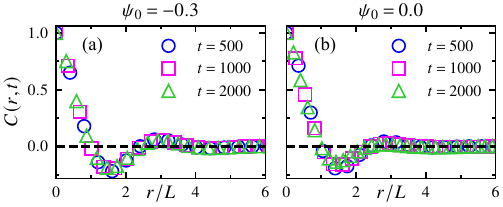}
    \caption{\label{fig:scaling} Scaled correlation function corresponding to the evolution shown in (a) Fig.~\ref{fig:snap_off_critical} and (b) Fig.~\ref{fig:snap_critical}, demonstrating data collapse at different times.}
\end{figure}

Having established that the model accurately captures the temporal evolution of the characteristic length scale, we now examine whether it also reproduces the underlying scaling behavior of the system. To this end, we analyze the two-point correlation function $C(r,t)$ at different times by rescaling the spatial coordinate with the characteristic domain size $L(t)$. Figure~\ref{fig:scaling} shows the scaled correlation function $C(r,t)$ plotted as a function of the reduced distance $r/L(t)$ for different times. For all compositions, the curves corresponding to different time instants collapse onto a single master curve, demonstrating the emergence of dynamical scaling during the coarsening process. This collapse indicates that the evolving domain morphology is statistically self-similar, with the characteristic length scale $L(t)$ serving as the only relevant length scale governing the structure of the system. The quality of the collapse remains consistent across both off-critical and critical compositions, as seen in Figs.~\ref{fig:scaling}(a) and (b), respectively. In general, the overall agreement confirms that the model successfully captures not only the growth kinetics but also the correct spatial correlations underlying the phase separation process.

These results, together with the agreement in $L(t)$, demonstrate that the proposed framework accurately reproduces the essential physical features of phase separation, including domain growth and dynamical scaling, over extended time evolution.

\section{\label{sec:conc} Conclusions}
We propose a physics-constrained neural surrogate to predict the domain growth in systems with conserved kinetics. The proposed framework is trained to capture domain coarsening in binary mixtures, where, following a sudden quench, the system evolves through the formation and subsequent growth of domains. In such systems, the overall composition remains conserved throughout the evolution. To achieve this, we imposed the conservation law directly on the network output as a hard constraint, so that the predicted composition is conserved to machine precision for any choice of network parameters. To assess the importance of this construction, we also trained a variant in which the projection is removed, and conservation is imposed only through a penalty term in the loss function. In this case, the composition drifts steadily during the autoregressive rollout, and the prediction accuracy degrades until $R^{2}$ becomes almost zero, whereas the hard-constrained model retains $R^{2}$ above $0.8$ over the same interval. Imposing the conservation law through the loss function is therefore not sufficient. It must be built into the network output to remain valid over long rollouts. 

Furthermore, the use of residual blocks enables the effective training of deeper networks. Additionally, attention mechanisms are incorporated at the lowest resolution to balance computational efficiency while accurately capturing long-range spatial correlations. The model is trained on early-time evolution snapshots for a system size of $64 \times 64$, with the objective of predicting a single snapshot at a time $100\Delta t$ ahead. The full spatiotemporal evolution is then generated by recursively feeding the model outputs back as inputs. Because the network is fully convolutional and uses circular padding throughout, the same trained model is applied without retraining to systems of size $L = 160$, $192$, and $256$, where it reproduces the coarsening dynamics with no noticeable loss of accuracy.

In this work, we evaluate the model performance when predictions are initiated from both later time snapshots and initial homogeneous conditions. Predictions starting from $t = 100$ show better one-to-one correspondence with the ground truth compared to those initiated from $t=0$. Notably, even in the absence of a strict one-to-one agreement, the model still captures the correct statistical and dynamical behavior over rollouts extending to $t = 2000$, well beyond the $t\in[0,300]$ window used for training. This is reflected in the accurate domain growth law obtained from the evolution of the characteristic length scale, which shows excellent agreement with the Lifshitz–Slyozov growth law $L(t) \sim t^{1/3}$. The surrogate prediction also exhibits dynamical scaling, evidenced by the collapse of the correlation function $C(r,t)$ when plotted against the rescaled distance $r/L(t)$. The present surrogate is, however, tied to a fixed prediction interval of $100\Delta t$, and extending it to larger prediction steps is a natural extension.

Finally, the model accurately learns the system dynamics while preserving the underlying conservation law by construction, enabling its extension beyond this prototypical problem to a broader class of systems with conserved kinetics such as surface-directed spinodal decomposition~\cite{Zaidi2022, Zaidi2024, Jaiswal2012}. This requires prior knowledge of which quantity is conserved, since the constraint is encoded in the network architecture rather than learned from data. The construction extends directly to any linearly conserved quantity, for which the projection of Eq.~\eqref{eq:proj} applies with the spatial mean replaced by the appropriate weighted inner product; systems with nonlinear invariants, for which no such closed-form projection exists, remain an open direction for future work.

\section*{ACKNOWLEDGEMENTS}
P.K.J. acknowledges the financial support from Anusandhan National Research Foundation (ANRF), India (Grant No. CRG/2022/006365) and IIT Jodhpur for a Seed Grant (No. I/SEED/PKJ/20220016).

\section*{AUTHOR DECLARATIONS}
\subsection*{Conflict of Interest}
All authors declare they have no competing interests.


\section*{DATA AVAILABILITY}
The code to generate the data and reproduce the results is available at \url{https://github.com/iitj-cosml/CH-UNET}.

\section*{References}
\nocite{*}
\bibliography{citations}

\end{document}